\RequirePackage{fix-cm}
\documentclass[twocolumn,final]{svjour3}     
\smartqed  
\usepackage{graphicx}
\usepackage[english]{babel}
\usepackage{appendix}
\usepackage{color,soul}
\usepackage{url}
\journalname{}
\begin{document}

\title{A Transformer-based approach to Irony and Sarcasm detection} 



\author{Rolandos Alexandros Potamias       \and
        Georgios Siolas \and Andreas - Georgios Stafylopatis 
}


\institute{Rolandos Alexandros Potamias$\dagger$ \at
                Department of Computing,\\
              Imperial College London, United Kingdom\\
              \email{r.potamias@imperial.ac.uk} \\  
              $\dagger$ Work performed while at National Technical University of Athens.
           \and
            Georgios Siolas \at
              School of Electrical and Computer Engineering,\\
              National Technical University of Athens, Greece\\
              \email{gsiolas@islab.ntua.gr}  
              \and
              Andreas - Georgios Stafylopatis \at
            School of Electrical and Computer Engineering, \\
            National Technical University of Athens, Greece\\
              \email{andreas@cs.ntua.gr} 
}

\date{Received: 12 November 2019 / Accepted: 4 June 2020}

\maketitle

\begin{abstract}
Figurative Language (FL) seems ubiquitous in all social-media discussion forums and chats, posing extra challenges to sentiment analysis endeavors. Identification of FL schemas in short texts remains largely an unresolved issue in the broader field of Natural Language Processing (NLP), mainly due to their contradictory and metaphorical meaning content. The main FL expression forms are sarcasm, irony and metaphor. In the present paper we employ advanced Deep Learning (DL) methodologies to tackle the problem of identifying the aforementioned FL forms.  Significantly extending our previous work \cite{potamias2019robust}, we propose a neural network methodology that builds on a recently proposed pre-trained transformer-based network architecture which, is further enhanced with the employment and devise of a recurrent convolutional neural network (RCNN). With this set-up, data preprocessing is kept in minimum. The performance of the devised hybrid neural architecture is tested on four benchmark datasets, and contrasted with other relevant state of the art methodologies and systems. Results demonstrate that the proposed methodology achieves state of the art performance under all benchmark datasets, outperforming, even by a large margin, all other methodologies and published studies.

\end{abstract}

\section{Introduction}
\label{intro}
In the networked-world era the production of (structured or unstructured) data is increasing with most of our knowledge being created and communicated via web-based social channels \cite{Winbey2019}. Such data explosion raises the need for efficient and reliable solutions for the management, analysis and interpretation of huge data sizes. Analyzing and extracting knowledge from massive data collections is not only a big issue per-se, but also challenges the data analytics state-of-the-art \cite{Zhou2017}, with statistical and machine learning methodologies paving the way, and deep learning (DL) taking over and presenting highly accurate solutions \cite{goodfellow_deep_2016}. Relevant applications in the field of social media cover a wide spectrum, from the categorization of major disasters \cite{Joseph2018} and the identification of suggestions \cite{potamias-etal-2019-ntua} to inducing users’ appeal to political parties \cite{Antonakaki2017}.

The raising of computational social science \cite{Lazer2009}, and mainly its social media dimension \cite{FM3993}, challenge contemporary computational linguistics and text-analytics endeavors. The challenge concerns the advancement of text analytics methodologies towards the transformation of unstructured excerpts into some kind of structured data via the identification of special passage characteristics, such as its emotional content (e.g., anger, joy, sadness) \cite{Kim2018}. In this context, Sentiment Analysis (SA) comes into play, targeting the devise and development of efficient algorithmic processes for the automatic extraction of a writer’s sentiment or emotion as conveyed in text excerpts. Relevant efforts focus on tracking the sentiment polarity of single utterances, which in most cases is loaded with a lot of subjectivity and a degree of vagueness \cite{DBLP:books/daglib/0036864}. Contemporary research in the field utilizes data from social media resources (e.g., Facebook, Twitter) as well as other short text references in blogs, forums etc \cite{potamias-siolas-2019-ntua}. However, users of social media tend to violate common grammar and vocabulary rules and even use various figurative language forms to communicate their message. In such situations, the sentiment inclination underlying the literal content of the conveyed concept may significantly differ from its figurative context, making SA tasks even more puzzling. Evidently, single turn text lack in detecting sentiment polarity on sarcastic and ironic expressions, as already signified in the relevant “SemEval-2014 Sentiment Analysis task 9” \cite{rosenthal_semeval-2014_2014-1}. Moreover, lacking of facial expressions and voice tone require context aware approaches to tackle such a challenging task and overcome its ambiguities \cite{Gupta2017}. As sentiment is the emotion behind customer engagement, SA finds its realization in automated customer aware services, elaborating over user’s emotional intensities \cite{Cuccio2014}. Most of the related studies utilize single turn texts from topic specific sources, such as Twitter, Amazon, IMDB etc. Hand crafted and sentiment-oriented features, indicative of emotion polarity, are utilized to represent respective excerpt cases. The formed data are then fed traditional machine learning classifiers (e.g. SVM, Random Forest, multilayer perceptrons) or DL techniques and respective complex neural architectures, in order to induce analytical models that are able to capture the underlying sentiment content and polarity of passages \cite{Hangya2017,Singh2018,Jianqiang2018}.

The linguistic phenomenon of figurative language (FL) refers to the contradiction between the literal and the non-literal meaning of an utterance \cite{devlin-etal-2019-bert}. Literal written language assigns ‘exact’ (or ‘real’) meaning to the used words (or phrases) without any reference to putative speech figures. In contrast, FL schemas exploit non-literal mentions that deviate from the exact concept presented by the used words and phrases. FL is rich of various linguistic phenomena like ‘metonymy’ reference to an entity stands for another of the same domain, a more general case of ‘synonymy’; and ‘metaphors’ systematic interchange between entities from different abstract domains \cite{dridi2019leveraging}. Besides the philosophical considerations, theories and debates about the exact nature of FL, findings from the neuroscience research domain present clear evidence on the presence of differentiating FL processing patterns in the human brain \cite{Weiland2014,cambridgehandbookpsycholinguistics:2012,Kasparian2013,Benedek2014,Cuccio2014}, even for woman-man attraction situations! \cite{Gao2017}. A fact that makes FL processing even more challenging and difficult to tackle. Indeed, this is the case of pragmatic FL phenomena like irony and sarcasm that main intention of in most of the cases, are characterized by an oppositeness to the literal language context. It is crucial to distinguish between the literal meaning of an expression considered as a whole from its constituents’ words and phrases. As literal meaning is assumed to be invariant in all context at least in its classical conceptualization \cite{Katz1977}, it is exactly this separation of an expression from its context that permits and opens the road to computational approaches in detecting and characterizing FL utterance.

We may identify three common FL expression forms namely, irony, sarcasm and metaphor. In this paper, figurative expressions, and especially ironic or sarcastic ones, are considered as a way of indirect denial. From this point of view, the interpretation and ultimately identification of the indirect meaning involved in a passage does not entail the cancellation of the indirectly rejected message and its replacement with the intentionally implied message (as advocated in \cite{Clark1984,Grice2008}). On the contrary ironic/sarcastic expressions presupposes the processing of both the indirectly rejected and the implied message so that the difference between them can be identified. This view differs from the assumption that irony and sarcasm involve only one interpretation \cite{w._gibbs_psycholinguistics_1986,sperber_irony_1981}. Holding that irony activates both grammatical / explicit as well as ironic / involved notions provides that irony will be more difficult to grasp than a non-ironic use of the same expression. 

Despite that all forms of FL are well studied linguistic phenomena \cite{w._gibbs_psycholinguistics_1986}, computational approaches fail to identify the polarity of them within a text. The influence of FL in sentiment classification emerged both on SemEval-2014 Sentiment Analysis task \cite{rosenthal_semeval-2014_2014-1} and \cite{dridi2019leveraging}. Results show that Natural Language Processing (NLP) systems effective in most other tasks see their performance drop when dealing with figurative forms of language. Thus, methods capable of detecting, separating and classifying forms of FL would be valuable building blocks for a system that could ultimately provide a full-spectrum sentiment analysis of natural language. 

In literature we encounter some major drawbacks of previous studies and we aim to resolve with our proposed method: 
\begin{itemize}
    \item[\textbullet ]Many studies tackle figurative language by utilizing a wide range of engineered features (e.g. lexical and sentiment based features) \cite{farias2016irony,gonzalez_identifying_2011,potamias2019robust,rajadesingan_sarcasm_2015,ravi_novel_2016,sulis2016figurative} making classification frameworks not feasible. 
    \item[\textbullet ]Several approaches search words on large dictionaries which demand large computational times and can be considered as impractical \cite{potamias2019robust,sulis2016figurative} 
  \item[\textbullet ] Many studies exhaustively preprocess the input texts, including stemming, tagging, emoji processing etc. that tend to be time consuming especially in large datasets \cite{kumar2019sarcasm,van2018exploring}. 
\item[\textbullet ] Many approaches attempt to create datasets using social media API’s to automatically collect data rather than exploiting their system on benchmark datasets, with proven quality. To this end, it is impossible to be compared and evaluated \cite{kumar2019sarcasm,ling2016empirical,van2018exploring}. 
\end{itemize}

To tackle the aforementioned problems, we propose an end-to-end methodology containing none hand crafted engineered features or lexicon dictionaries, a preprocessing step that includes only de-capitalization and we evaluate our system on several benchmark dataset. To the best of our knowledge, this is the first time that an unsupervised pre-trained Transformer method is used to capture figurative language in many of its forms. 

The rest of the paper is structured as follows, in Section \ref{sec:1} we present the related work on the field of FL detection, in Section \ref{sec:2} we shortly describe the background of recent advances in natural language processing that achieve high performance in a wide range of tasks and will be used to compare performance, in \ref{sec:3} we present our proposed method, the results of our experiments are presented in Section \ref{sec:3}, and finally our conclusion is in Section \ref{sec:5}. 
 
\section{Literature Review}
\label{sec:1}
Although the NLP community have researched all aspects of FL independently, none of the proposed systems were evaluated on more than one type. Related work on FL detection and classification tasks could be categorized into two main categories, according to the studied task: (a) irony and sarcasm detection, and (b) sentiment analysis of FL excerpts. Even if sarcasm and irony are not identical phenomenons, we will present those types together, as they appear in the literature.  

\subsection{Irony and Sarcasm Detection} 
Recently, the detection of ironic and sarcastic meanings from respective literal ones have raised scientific interest due to the intrinsic difficulties to differentiate between them. Apart from English language, irony and sarcasm detection have been widely explored on other languages as well, such as Italian \cite{stranisci2016annotating}, Japanese \cite{hiai2018sarcasm}, Spanish \cite{ortega2019overview}, Greek \cite{charalampakis2016comparison} etc. In the review analysis that follows we group related approaches according to the their adopted key concepts to handle FL. 
\medskip

\textbf{Approaches based on unexpectedness and contradictory factors.} Reyes et al. \cite{reyes_humor_2012,reyes_multidimensional_2013} were the first that attempted to capture irony and sarcasm in social media. They introduced the concepts of unexpectedness and contradiction that seems to be frequent in FL expressions. The unexpectedness factor was also adopted as a key concept in other studies as well. In particular, Barbieri et al. \cite{barbieri_modelling_2014} compared tweets with sarcastic content with other topics such as, \#politics, \#education, \#humor. The measure of unexpectedness was calculated using the \textit{American National Corpus Frequency Data} source as well as the morphology of tweets, using Random Forests (RF) and Decision Trees (DT) classifiers. In the same direction, Buschmeir et al. \cite{buschmeier_impact_2014} considered unexpectedness as an emotional imbalance between words in the text. Ghosh et al. \cite{ghosh_sarcastic_2015} identified sarcasm using Support Vector Machines (SVM) using as features the identified contradictions within each tweet. 
\medskip

\textbf{Content and context-based approaches.}
Inspired by the contradictory and unexpectedness concepts, follow-up approaches utilized features that expose information about the content of each passage including: N-gram patterns, acronyms and adverbs \cite{carvalho_clues_2009}; semi-supervised attributes like word frequencies \cite{davidov_semi-supervised_2010}; statistical and semantic features \cite{ravi_novel_2016}; and \textit{Linguistic Inquiry and Word Count} (LIWC) dictionary along with syntactic and psycho-linguistic features \cite{radford2018improving}. LIWC corpus \cite{pennebaker_linguistic_1999} was also utilized in \cite{gonzalez_identifying_2011}, comparing sarcastic tweets with positive and negative ones using an SVM classifier. Similarly, using several lexical resources \cite{sulis2016figurative}, and syntactic and sentiment related features \cite{ling2016empirical}, the respective researchers explored differences between sarcastic and ironic expressions. Affective and structural features are also employed to predict irony with conventional machine learning classifiers (DT, SVM, Naïve Bayes/NB) in \cite{farias2018knowledge}. In a follow-up study \cite{farias2016irony}, a knowledge-based k-NN classifier was fed with a feature set that captures a wide range of linguistic phenomena (e.g., structural, emotional). Significant results were achieved in \cite{van2018exploring}, were a combination of lexical, semantic and syntactic features passed through an SVM classifier that outperformed LSTM deep neural network approaches. Apart from local content, several approaches claimed that global context may be essential to capture FL phenomena. In particular, in \cite{wallace_sparse_2015} it is claimed that capturing previous and following comments on Reddit increases classification performance. Users’ behavioral information seems to be also beneficial as it captures useful contextual information in Twitter post \cite{rajadesingan_sarcasm_2015}. A novel unsupervised probabilistic modeling approach to detect irony was also introduced in \cite{nozza2016unsupervised}.
\medskip

\textbf{Deep Learning approaches.} Although several DL methodologies, such as recurrent neural networks (RNNs), are able to capture hidden dependencies between terms within text passages and can be considered as content-based, we grouped all DL studies for readability purposes. Word Embeddings, i.e., learned mappings of words to real valued vectors \cite{mikolov_efficient_2013}, play a key role in the success of RNNs and other DL neural architectures that utilize pre-trained word embeddings to tackle FL. In fact, the combination of word embeddings with Convolutional Neural Networks (CNN), so called CNN-LSTM units, was introduced by Kumar \cite{kumar_having_2017} and Ghosh \& Veale \cite{ghosh_fracking_2016} achieving state-of-the-art performance. Attentive RNNs exhibit also good performance when matched with pre-trained Word2Vec embeddings \cite{huang_irony_2017}, and contextual information \cite{zhang2019irony}. Following the same approach an LSTM based intra-attention was introduced in \cite{tay-etal-2018-reasoning} that achieved increased performance. A different approach, founded on the claim that number present significant indicators, was introduced by Dubey et al. \cite{dubey2019numbers}. Using an attentive CNN on a dataset with sarcastic tweets that contain numbers, showed notable results. An ensemble of a shallow classifier with lexical, pragmatic and semantic features, utilizing a Bidirectional LSTM model is presented in \cite{kumar2019empirical}. In a subsequent study \cite{kumar2019sarcasm}, the researchers engineered a soft attention LSTM model coupled with a CNN. Contextual DL approaches are also employed, utilizing pre-trained along with user embeddings structured from previous posts \cite{amir2016modelling} or, personality embeddings passed through CNNs \cite{hazarika2018cascade}. ELMo embeddings \cite{peters2018deep} are utilized in \cite{ilic2018deep}. In our previous approach we implemented an ensemble deep learning classifier (DESC) \cite{potamias2019robust}, capturing content and semantic information. In particular, we employed an extensive feature set of a total 44 features leveraging syntactic, demonstrative, sentiment and readability information from each text along with Tf-idf features. In addition, an attentive bidirectional LSTM model trained with GloVe pre-trained word embeddings was utilized to structure an ensemble classifier processing different text representations. DESC model performed state-of-the-art results on several FL tasks. 

\subsection{Sentiment Analysis on Figurative Language}
The Semantic Evaluation Workshop-2015 \cite{ghosh_semeval-2015_2015} proposed a joint task to evaluate the impact of FL in sentiment analysis on ironic, sarcastic and metaphorical tweets, with a number of submissions achieving highly performance results.  The ClaC team \cite{ozdemir_clac-sentipipe:_2015} exploited four lexicons to extract attributes as well as syntactic features to identify sentiment polarity. The UPF team \cite{barbieri_upf-taln:_2015} introduced a regression classification methodology on tweet features extracted with the use of the widely utilized SentiWordNet and DepecheMood lexicons. The LLT-PolyU team \cite{xu_llt-polyu:_2015} used semi-supervised regression and decision trees on extracted uni-gram and bi-gram features, coupled with features that capture potential contradictions at short distances. An SVM-based classifier on extracted n-gram and Tf-idf features was used by the Elirf team \cite{gimenez_elirf:_2015} coupled with specific lexicons such as Affin, Patter and Jeffrey 10. Finally, the LT3 team \cite{van_hee_lt3:_2015} used an ensemble Regression and SVM semi-supervised classifier with lexical features extracted with the use of WordNet and DBpedia11.

\section{The background: Recent advances in Natural Language Processing }
\label{sec:2}
Due to the limitations of annotated datasets and the high cost of data collection, unsupervised learning approaches tend to be an easier way towards training networks. Recently, \textit{transfer learning} approaches, i.e., the transfer of already acquired knowledge to new conditions, are gaining attention in several domain adaptation problems \cite{ganin2016domain}. In fact, pre-trained embeddings representations, such as GloVe, ElMo and USE, coupled with transfer learning architectures were introduced and managed to achieve state-of-the-art results on various NLP tasks \cite{howard-ruder-2018-universal}. In the current section we summarize those methods in order to introduce our proposed transfer learning system in Section \ref{sec:4}. Model specifications used for the state-of-the-art models can be found in Appendix \ref{appendix}. 

\subsection{Contextual Embeddings}
Pre-trained word embeddings proved to increase classification performances in many NLP tasks. In particular, Global Vectors (GloVe) \cite{pennington_glove:_2014} and Word2Vec \cite{mikolov_distributed_2013} became popular in various tasks due to their ability to capture representative semantic representations of words, trained on large amount of data. However, in various studies (e.g., \cite{peters2017semi,peters2018deep,mccann2017learned}) it is argued that the actual meaning of words along with their semantics representations varies according to their context. Following this assumption, researchers in \cite{peters2018deep} present an approach that is based on the creation of pre-trained word embeddings through building a bidirectional Language model, i.e. predicting next word within a sequence. The ELMo model was exhaustingly trained on 30 million sentences corpus \cite{chelba2013one}, with a two layered bidirectional LSTM architecture, aiming to predict both next and previous words, introducing the concept of contextual embeddings. The final embeddings vector is produced by a task specific weighted sum of the two directional hidden layers of LSTM models. Another contextual approach for creating embedding vector representations is proposed in \cite{cer2018universal} where, complete sentences, instead of words, are mapped to a latent vector space. The approach provides two variations of Universal Sentence Encoder (USE) with some trade-offs in computation and accuracy. The first approach consists of a computationally intensive transformer that resembles a transformer network \cite{vaswani2017attention}, proved to achieve higher performance figures. In contrast, the second approach provides a light-weight model that averages input embedding weights for words and bi-grams by utilizing of a Deep Average Network (DAN) \cite{iyyer-etal-2015-deep}. The output of the DAN is passed through a feedforward neural network in order to produce the sentence embeddings. Both approaches take as input lowercased PTB tokenized\footnote{\url{https://nlp.stanford.edu/software/tokenizer.html}} strings, and output a 512-dimensional sentence embedding vectors. 
\subsection{Transformer Methods}
Sequence-to-sequence (seq2seq) methods using encoder-decoder schemes are a popular choice for several tasks such as Machine Translation, Text Summarization, Question Answering etc. \cite{sutskever2014sequence}. However, encoder’s contextual representations are uncertain when dealing with long-range dependencies. To address these drawbacks, Vaswani et al. in \cite{vaswani2017attention} introduced a novel network architecture, called Transformer, relying entirely on self-attention units to map input sequences to output sequences without the use of RNNs. The Transformer’s decoder unit architecture contains a masked multi-head attention layer followed by a multi-head attention unit and a feed forward network whereas the decoder unit is almost identical without the masked attention unit. Multi-head self-attention layers are calculated in parallel facing the computational costs of regular attention layers used by previous seq2seq network architectures. 
In \cite{devlin-etal-2019-bert} the authors presented a model that is founded on findings from various previous studies (e.g., \cite{dai2015semi,howard2018universal,peters2018deep,radford2018improving,vaswani2017attention}), which achieved state-of-the-art results on eleven NLP tasks, called BERT - Bidirectional Encoder Representations from Transformers. The BERT training process is split in two phases, the unsupervised pre-training phase and the fine-tuning phase using labelled data for down-streaming tasks. In contrast with previous proposed models (e.g., \cite{peters2018deep,radford2018improving}), BERT uses masked language models (MLMs) to enable pre-trained deep bidirectional representations. In the pre-training phase the model is trained with a large amount of unlabeled data from Wikipedia, BookCorpus \cite{zhu2015aligning} and WordPiece \cite{wu2016google} embeddings. In this training part, the model was tested on two tasks; on the first task, the model randomly masks 15\% of the input tokens aiming to capture conceptual representations of word sequences by predicting masked words inside the corpus, whereas in the second task the model is given two sentences and tries to predict whether the second sentence is the next sentence of the first. In the second phase, BERT is extended with a task-related classifier model that is trained on a supervised manner. During this supervised phase, the pre-trained BERT model receives minimal changes, with the classifier’s parameters trained in order to minimize the loss function. Two models presented in \cite{devlin-etal-2019-bert}, a “Base Bert” model with 12 encoder layers (i.e. transformer blocks), feed-forward networks with 768 hidden units and 12 attention heads, and a “Large Bert” model with 24 encoder layers 1024 feed-the pre-trained Bert model, an architecture almost identical with the aforementioned Transformer network. A [CLS] token is supplied in the input as the first token, the final hidden state of which is aggregated for classification tasks. Despite the achieved breakthroughs, the BERT model suffers from several drawbacks. Firstly, BERT, as all language models using Transformers, assumes (and pre-supposes) independence between the masked words from the input sequence, and neglects all the positional and dependency information between words. In other words, for the prediction of a masked token both word and position embeddings are masked out, even if positional information is a key-aspect of NLP \cite{dai2019transformer}. In addition, the [MASK] token which, is substituted with masked words, is mostly absent in fine-tuning phase for down-streaming tasks, leading to a pre-training fine-turning discrepancy. To address the cons of BERT, a permutation language model was introduced, so-called XLnet, trained to predict masked tokens in a non-sequential random order, factorizing likelihood in an autoregressive manner without the independence assumption and without relying on any input corruption \cite{yang2019xlnet}. In particular, a query stream is used that extends embedding representations to incorporate positional information about the masked words. The original representation set (content stream), including both token and positional embeddings, is then used as input to the query stream following a scheme called “Two-Stream SelfAttention”. To overcome the problem of slow convergence the authors propose the prediction of the last token in the permutation phase, instead of predicting the entire sequence. Finally, XLnet uses also a special token for the classification and separation of the input sequence, [CLS] and [SEP] respectively, however it also learns an embedding that denotes whether the two words are from the same segment. This is similar to relative positional encodings introduced in TrasformerXL \cite{dai2019transformer}, and extents the ability of XLnet to cope with tasks that encompass arbitrary input segments. Recently, a replication study, \cite{liu2019roberta}, suggested several modifications in the training procedure of BERT which, outperforms the original XLNet architecture on several NLP tasks. The optimized model, called Robustly Optimized BERT Approach (RoBERTa), used 10 times more data (160GB compared with the 16GB originally exploited), and is trained with far more epochs than the BERT model (500K vs 100K), using also 8-times larger batch sizes, and a byte-level BPE vocabulary instead of the character-level vocabulary that was previously utilized. Another significant modification, was the dynamic masking technique instead of the single static mask used in BERT. In addition, RoBERTa model removes the next sentence prediction objective used in BERT, following advises by several other studies that question the NSP loss term \cite{lample2019cross,you2019reducing,joshi2019spanbert}.

\section{Proposed Method:  Recurrent CNN RoBERTA (RCNN-RoBERTa)} 
\label{sec:3}
\begin{figure}
\centering
  \includegraphics[scale=0.65]{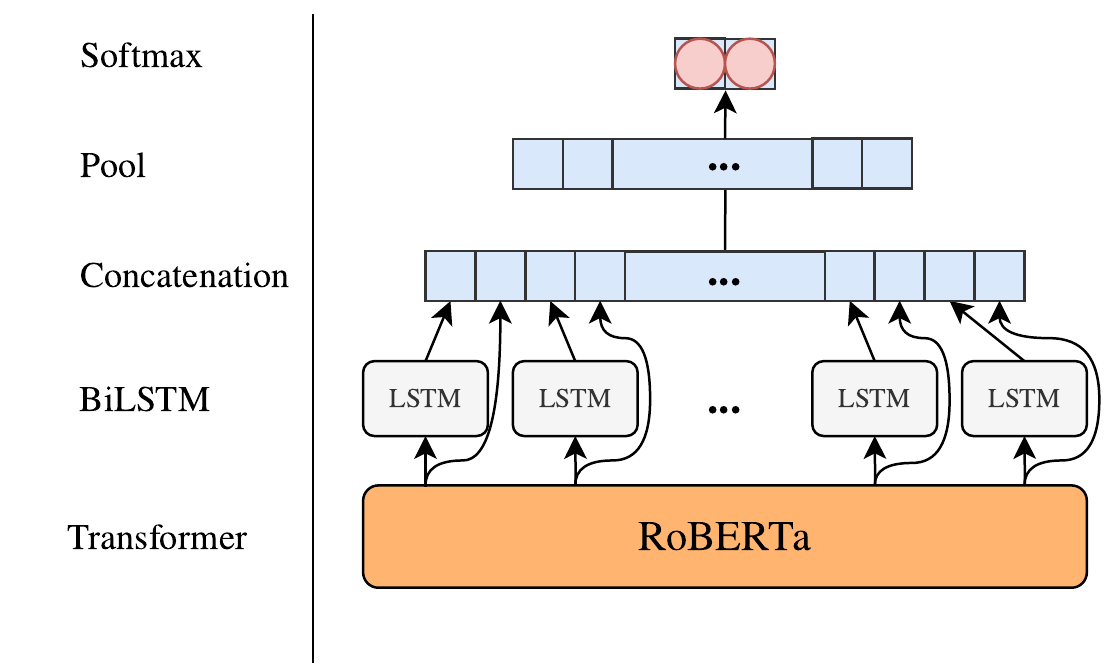}
\caption{The proposed RCNN-RoBERTa methodology, consisting of a RoBERTa pre-trained transformer followed by a Bidirectional LSTM layer (BiLSTM). Pooling is applied to the representation vector of concatenated RoBERTa and LSTM outputs and passed through a fully connected softmax-activated layer.  We refer the reader to \cite{liu2019roberta,vaswani2017attention} for RoBERTa Transformer-based architecture.}
\label{fig:1}       
\end{figure}

The intuition behind our proposed RCNN-RoBERTa approach is founded on the following observation: as pre-trained networks are beneficial for several down-streaming tasks, their outputs could be further enhanced if processed properly by other networks. Towards this end, we devised an end-to-end model that utilizes pre-trained RoBERTa \cite{liu2019roberta} weights combined with a RCNN in order to capture contextual information. The RoBERTa network architecture is utilized in order to efficiently map words onto a rich embedding space. To improve RoBERTa’s performance and identify FL within a sentence, it is essential to capture the dependencies within RoBERTa’s pre-trained word-embeddings. This task can be tackled with an RNN layer suited to capture temporal reliant information, in contrast, to fully-connected and 1D convolution layers that are not able to delineate with such dependencies. In addition, aiming to enhance the proposed network architecture, the RNN layer is followed with a fully connected layer that simulates 1D convolution with a large kernel (see below), which is capable to capture spatio-temporal dependencies in RoBERTa’s projected latent space. Actually, the proposed leaning model is based on a hybrid DL neural architecture that utilizes pre-trained transformer models and feed the hidden representations of the transformer into a Recurrent Convolutional Neural Network (RCNN), similar to \cite{lai2015recurrent}. In particular, we employed the RoBERTa base model with 12 hidden states and 12 attention heads, and used its output hidden states as an embedding layer to a RCNN. As already stated, contradictions and long-time dependencies within a sentence may be used as strong identifiers of FL expressions. RNNs are often used to capture temporal relationships between words. However they are strongly biased, i.e. later words are tending to be more dominant that previous ones. This problem can be alleviated with CNNs, which, as unbiased models, can determine semantic relationships between words with max-pooling \cite{lai2015recurrent,nguyen_relation}. Nevertheless, contextual information in CNNs is depended totally on kernel sizes. Thus, we appropriately modified the RCNN model presented in \cite{lai2015recurrent} in order to capture unbiased recurrent informative relationships within text. In particular, we implemented a Bidirectional LSTM (BiLSTM) layer, which is fed with RoBERTa’s final hidden layer weights. The output of LSTM is concatenated with the embedded weights, and passed through a feedforward network, acting as a 1D convolution layer with large kernel, and a max-pooling layer. Finally, softmax function is used for the output layer. Table \ref{hyper} shows the parameters used in training and Figure \ref{fig:1}
illustrates the proposed deep network architecture. 
\begin{table}[]
\centering
\caption{Selected hyperparameters used in our proposed method RCNN-RoBERTa. The hyperparameters where settled following a grid search based on a 5-fold cross-validation process; the finally selected parameters are the ones that exhibit the best performance.}
\begin{tabular}{|l|c|}
\hline
\textit{Hyperparameter}    & Value \\ \hline \hline
RoBERTa Layers          & 12    \\ \hline
RoBERTa Attention Heads & 12    \\ \hline
LSTM units             & 64    \\ \hline
LSTM dropout           & 0.1   \\ \hline
Batch size             & 10    \\ \hline
Adam epsilon           & 1e-6  \\ \hline
Epochs                 & 5     \\ \hline
Learning rate          & 2e-5  \\ \hline
Weight decay           & 1e-5  \\ \hline \hline
\end{tabular}

\label{hyper}

\end{table}

\section{Experimental Results} 
\label{sec:4}
To assess the performance of the proposed method we performed an exhaustive comparison with several advanced state-of-the-art methodologies along with published results. Nowadays trends in NLP community tend to explicitly utilize deep learning methodologies as the most convenient way to approach various semantic analysis tasks. In the past decade, RNNs such as LSTM and GRUs were the most popular choice, whereas the last years the impact of attention-based models such as Transformers seems to outperform all previous methods, even by a large margin \cite{vaswani2017attention,devlin-etal-2019-bert}. On the contrary, classical machine learning algorithms such as SVM, k-Nearest Neighbors (kNN) and tree-based models (Decision Trees, Random Forest) have been considered inappropriate for real world applications, due to their demand on hand-crafted feature extraction and exhaustive preprocessing strategies. In order to have a reasonable kNN or SVM algorithm, there should be a lot of effort to embed sentences on word level to a higher space that a classifier may recognize patterns. In support of the arguments made, in our previous study \cite{potamias2019robust}, classical machine learning algorithms supported with rich and informative features failed to compete deep learning methodologies and proved non-feasible to FL detection. To this end, in this study we acquired several state-of-the-art models to compare our proposed method. The used methodologies were appropriately implemented using the available codes and guidelines, and include: ELMo \cite{peters2018deep}, USE \cite{cer2018universal}, NBSVM \cite{wang2012baselines}, FastText \cite{joulin2016fasttext},  XLnet base cased model (XLnet) \cite{yang2019xlnet}, BERT \cite{devlin-etal-2019-bert} in two setups: BERT base cased (BERT-Cased) and BERT base uncased (BERT-Uncased) models, and RoBERTa base model \cite{liu2019roberta}. The settings and the hyper-parameters used for training the aforementioned models can be found in Appendix \ref{appendix}. The published results were acquired from the respective original publication (the reference publication is indicated in the respective tables). For the comparison we utilized benchmark datasets that include ironic, sarcastic and metaphoric expressions. Namely, we used the dataset provided in “Semantic Evaluation Workshop Task 3” (SemEval-2018) that contains ironic tweets \cite{hee_semeval-2018_2018}; Riloff’s high quality sarcastic unbalanced dataset \cite{riloff_sarcasm_2013}; a large dataset containing political comments from Reddit \cite{khodak_large_2017}; and a SA dataset that contains tweets with various FL forms from “SemEval-2015 Task 11” \cite{ghosh_semeval-2015_2015}. All datasets are used in a binary classification manner (i.e., irony/sarcasm vs. literal), except from the “SemEval-2015 Task 11” dataset where the task is to predict a sentiment integer score (from -5 to 5) for each tweet (refer to \cite{potamias2019robust} for more details). For a fair comparison, we splitted the datasets on train/test stets as proposed by the authors providing the datasets or by following the settings of the respective published studies. The evaluation was made across standard five metrics namely, Accuracy (Acc), Precision (Pre), Recall (Rec), F1-score (F1), and Area Under the Receiver Operating Characteristics Curve (AUC). For the SA task the cosine similarity metric (Cos) and mean squared error (MSE) metrics are used, as proposed in the original study \cite{ghosh_semeval-2015_2015}. 

The results are summarized in the tables \ref{TableIrony}-\ref{TableFL}; each table refers to the respective comparison study. All tables present the performance results of our proposed method (“Proposed”) and contrast them to eight state-of-the-art baseline methodologies along with published results using the same dataset. Specifically, Table \ref{TableIrony} presents the results obtained using the ironic dataset used in SemEval-2018 Task 3.A, compared with recently published studies and two high performing teams from the respective SemEval shared task \cite{baziotis_ntua-slp_2018,wu_thu_ngn_2018}. Tables \ref{TableReddit},\ref{Tableriloff} summarize results obtained using Sarcastic datasets (Reddit SARC politics \cite{khodak_large_2017} and Riloff Twitter \cite{riloff_sarcasm_2013}). Finally, Table \ref{TableFL} compares the results from baseline models, from top two ranked task participants \cite{barbieri_upf-taln:_2015,ozdemir_clac-sentipipe:_2015}, from our previous study with the DESC methodology \cite{potamias2019robust} with the proposed RCNN-RoBERTa framework on a Sentiment Analysis task with figurative language, using the SemEval 2015 Task 11 dataset. 

\begin{table}[]
\caption{Comparison of RCNN-RoBERTa with state-of-the-art neural network classifiers and published results on SemEval-2018 dataset; bold figures indicate superior performance.}
\centering
\begin{tabular}{|l|c|c|c|c|c|}
\hline
\multicolumn{6}{|c|}{\textbf{Irony/SemVal-2018-Task 3.A}\cite{hee_semeval-2018_2018}}                                                                                     \\ \hline
\textit{System}   & \textbf{Acc}  & \textbf{Pre}  & \textbf{Rec}  & \textbf{F1}   & \textbf{AUC}  \\ \hline \hline
ELMo              & 0.66                  & 0.66                  & 0.67                  & 0.66                 & 0.72                  \\ \hline
USE               & 0.69                  & 0.67                  & 0.67                  & 0.67                 & 0.74                  \\ \hline
NBSVM             & 0.69                  & 0.70                  & 0.69                  & 0.69                 & 0.73                  \\ \hline
FastText          & 0.69                  & 0.71                  & 0.69                  & 0.69                 & 0.73                  \\ \hline
XLnet   & 0.71                  & 0.71                  & 0.72                  & 0.70                 & 0.80                 \\ \hline
BERT-Cased   & 0.70                  & 0.69                  & 0.70                  & 0.69                 & 0.77                  \\ \hline
BERT-Uncased  & 0.69                  & 0.68                  & 0.69                  & 0.68                 & 0.77                  \\ \hline
RoBERTa    & 0.79                  & 0.78                  & 0.79                  & 0.78                 & 0.89                  \\ \hline \hline
Wu et al.\cite{wu_thu_ngn_2018}        & 0.74                  & 0.63                  & 0.80                  & 0.71                 & -                     \\ \hline
Ilić et al. \cite{ilic2018deep}             & 0.71                  & 0.70                  & 0.70                  & 0.70                 & -                     \\ \hline
THU\_NGN \cite{wu_thu_ngn_2018}        & 0.73                  & 0.63                  & 0.80                  & 0.71                 & -                     \\ \hline
NTUA-SLP \cite{baziotis_ntua-slp_2018}         & 0.73                  & 0.65                  & 0.69                  & 0.67                 & -                     \\ \hline
Zhang et al. \cite{zhang2019irony}            & -                     & -                     & -                     & 0.71                 & -                     \\ \hline
DESC \cite{potamias2019robust}              & 0.74                  & 0.73                  & 0.73                  & 0.73                 & 0.78                    \\ \hline
\hline
\textbf{Proposed} & \textbf{0.82}         & \textbf{0.81}         & \textbf{0.80}         & \textbf{0.80}        & \textbf{0.89}         \\ \hline
\end{tabular}
\label{TableIrony}
\end{table}

\begin{table}[]
\caption{Comparison of RCNN-RoBERTa with state-of-the-art neural network classifiers and published results on Reddit Politics dataset.}
\centering
\begin{tabular}{|l|c|c|c|c|c|}
\hline
\multicolumn{6}{|c|}{\textbf{Reddit SARC2.0 politics} \cite{khodak_large_2017}}                                                                                         \\ \hline
\textit{System}   & \textbf{Acc}  & \textbf{Pre}  & \textbf{Rec}  & \textbf{F1}   & \textbf{AUC}  \\ \hline \hline

ELMo              & 0.70          & 0.70          & 0.70          & 0.70          & 0.77          \\ \hline
USE               & 0.75          & 0.75          & 0.75          & 0.75          & 0.82          \\ \hline
NBSVM             & 0.65          & 0.65          & 0.65          & 0.65          & 0.68          \\ \hline
FastText          & 0.63          & 0.65          & 0.61          & 0.63          & 0.64          \\ \hline
XLnet   & 0.76          & 0.77          & 0.74          & 0.76          & 0.83          \\ \hline
BERT-Cased   & 0.76          & 0.76          & 0.75          & 0.76          & 0.84          \\ \hline
BERT-Uncased  & 0.77          & 0.77          & 0.77          & 0.77          & 0.84          \\ \hline
RoBERTa    & 0.77          & 0.77          & 0.77          & 0.77          & 0.85          \\ \hline \hline
CASCADE \cite{hazarika2018cascade}          & 0.74          & -             & -             & 0.75          & -             \\ \hline
Ilić et al. \cite{ilic2018deep}          & 0.79         & -             & -             & -             & -             \\ \hline
Khodak et al. \cite{khodak_large_2017}         & 0.77         & -             & -             & -             & -             \\ \hline \hline
\textbf{Proposed} & \textbf{0.79} & \textbf{0.78} & \textbf{0.78} & \textbf{0.78} & \textbf{0.85} \\ \hline
\end{tabular}
\label{TableReddit}
\end{table}

\begin{table}[]
\caption{Comparison of RCNN-RoBERTa with state-of-the-art neural network classifiers and published results on on Sarcastic Rillof’s dataset.}
\centering
\begin{tabular}{|l|c|c|c|c|c|}
\hline
\multicolumn{6}{|c|}{\textbf{Riloff Sarcastic Dataset}\cite{riloff_sarcasm_2013}}                                           \\ \hline
\textit{System}   & Acc           & Pre           & Rec           & F1            & AUC           \\ \hline \hline
ELMo              & 0.85          & 0.85          & 0.86          & 0.85          & 0.89          \\ \hline
USE               & 0.87          & 0.81          & 0.76          & 0.78          & 0.89          \\ \hline
NBSVM             & 0.75          & 0.59          & 0.57          & 0.58          & 0.60          \\ \hline
FastText          & 0.83          & 0.83          & 0.61          & 0.64          & 0.85          \\ \hline
XLnet   & 0.86          & 0.88          & 0.86          & 0.86          & 0.92          \\ \hline
BERT-Cased   & 0.86          & 0.87          & 0.85          & 0.86          & 0.91          \\ \hline
BERT-Uncased  & 0.87          & 0.88          & 0.87          & 0.87          & 0.91          \\ \hline
RoBERTa    & 0.89          & 0.85          & 0.84          & 0.85          & 0.91          \\ \hline \hline
Farrias et al. \cite{farias2018knowledge}   & -             & -             & -             & 0.75          & -             \\ \hline
Ilić et al. \cite{ilic2018deep}              & 0.86          & 0.78          & 0.77          & 0.75          & -             \\ \hline
Tay el at. \cite{tay-etal-2018-reasoning}               & 0.82          & 0.74          & 0.73          & 0.73          & -             \\ \hline
DESC \cite{potamias2019robust}             & 0.87          & 0.86          & 0.87          & 0.87          & 0.86            \\ \hline
Ghosh \cite{ghosh_fracking_2016}            & -             & 0.88          & 0.88          & 0.88          & -             \\ \hline \hline
\textbf{Proposed} & \textbf{0.91} & \textbf{0.90} & \textbf{0.90} & \textbf{0.90} & \textbf{0.94} \\ \hline

\end{tabular}
\label{Tableriloff}
\end{table}

\begin{table}[]
\caption{Comparison of RCNN-RoBERTa with state-of-the-art neural network classifiers and published results on Task11 - SemEval-2015 dataset (sentiment analysis of figurative language expression).}
\centering
\begin{tabular}{|l|c|c|}
\hline
\multicolumn{3}{|c|}{\textbf{SemEval-2015 Task 11 }\cite{ghosh_semeval-2015_2015}} \\ \hline
\textit{System}  & \textbf{COS}  & \textbf{MSE}  \\ \hline
ELMo             & 0.710         & 3.610         \\ \hline
USE              & 0.71          & 3.17          \\ \hline
NBSVM            & 0.69          & 3.23          \\ \hline
FastText         & 0.72          & 2.99          \\ \hline
XLnet  & 0.76          & 1.84          \\ \hline
BERT-Cased  & 0.72          & 1.97          \\ \hline
BERT-Uncased & 0.79          & 1.54          \\ \hline
RoBERTa   & 0.78          & 1.55          \\ \hline \hline
UPF  \cite{barbieri_upf-taln:_2015}           & 0.71          & 2.46          \\ \hline
ClaC \cite{ozdemir_clac-sentipipe:_2015}             & 0.76          & 2.12          \\ \hline
DESC  \cite{potamias2019robust}           & \textbf{0.82} & 2.48          \\ \hline \hline
\textbf{Proposed}         & 0.81          & \textbf{1.45} \\ \hline 
\end{tabular}
\label{TableFL}
\end{table}

As it can be easily observed, the proposed RCNN-RoBERTa approach outperforms all approaches as well as all methods with published results, for the respective binary classification tasks (Tables \ref{TableIrony}, \ref{TableReddit}, and \ref{Tableriloff}). In particular, the RCNN architecture seems to reinforce RoBERTa model by 2-5\% F1 score, increasing also the classification confidence, in terms of AUC performance. Note also that RoBERTa-RCNN show better behaviour, compared to RoBERTa, on imbalanced datasets (Riloff \cite{riloff_sarcasm_2013}, SemEval-2015 \cite{ghosh_semeval-2015_2015}). Also, one-way ANOVA Tukey test \cite{montgomery2017design} revealed that RoBERTa-RCNN model outperforms by a statistical significant margin the maximum values of all metrics of previously published approaches, i.e. $p=0.015; p<0.05$ for Ironic tweets and $p=0.003; p<0.01$ for Riloff Sarcastic tweets. Furthermore, the proposed method increased the state-of-the-art performance even by a large margin in terms of Accuracy, F1 and AUC score. Our previous approach, DESC (introduced in  \cite{potamias2019robust}), performs slightly better in terms of cosine similarity for the sentiment scoring task (Table \ref{TableFL}, 0,820 vs. 0,810), with the RCNN-RoBERTa approach to perform better and managing to significantly improve the MSE measure by almost 33.5\% (2,480 vs. 1,450).

\section{Conclusion} 
\label{sec:5}
In this study, we propose the first transformer based methodology, leveraging the pre-trained RoBERTa model combined with a recurrent convolutional neural network, to tackle figurative language in social media. Our network is compared with all, to the best of our knowledge, published approaches under four different benchmark dataset. In addition, we aim to minimize preprocessing and engineered feature extraction steps which are, as we claim, unnecessary when using overly trained deep learning methods such as transformers. In fact, hand crafted features along with preprocessing techniques such as stemming and tagging on huge datasets containing thousands of samples are almost prohibited in terms of their computation cost. 
Our proposed model, RCNN-RoBERTa, achieve state-of-the-art performance under six metrics over four benchmark dataset, denoting that transfer learning non-literal forms of language. Moreover, RCNN-RoBERTa model outperforms all other state-of-the-art approaches tested including BERT, XLnet, ELMo, and USE under all metric, some by a large factor.

%

\bibliographystyle{spmpsci}      
\bibliography{References2.bib}   

\appendix
\section{Appendix}
\label{appendix}
In our experiments we compared our model with several seven different classifiers under different settings. For the ELMo system we used the mean-pooling of all contextualized word representations, i.e. character-based embedding representations and the output of the two layer LSTM resulting with a 1024 dimensional vector, and passed it through two deep dense ReLu activated layers with 256 and 64 units. Similarly, USE embeddings are trained with a Transformer encoder and output 512 dimensional vector for each sample, which is also passed through through two deep dense ReLu activated layers with 256 and 64 units. Both ELMo and USE embeddings retrieved from TensorFlow Hub\footnote{\url{https://tfhub.dev/s?module-type=text-embedding}}. NBSVM system was modified according to \cite{wang2012baselines} and trained with a ${10^{-3}}$ leaning rate for 5 epochs with Adam optimizer \cite{kingma_adam:_2014}. FastText system was implemented by utilizing pre-trained embeddings \cite{joulin2016fasttext} passed through a global max-pooling and a 64 unit fully connected layer. System was trained with Adam optimizer with learning rate ${0.1}$ for 3 epochs. XLnet model implemented using the base-cased model with 12 layers, 768 hidden units and 12 attention heads. Model trained with learning rate ${4 \times 10^{-5}}$ using ${10^{-5}}$ weight decay for 3 epochs. We exploited both cased and uncased BERT-base models containing 12 layers, 768 hidden units and 12 attention heads. We trained models for 3 epochs with learning rate ${2 \times 10^{-5}}$ using ${10^{-5}}$ weight decay. We trained RoBERTa model following the setting of BERT model. RoBERTa, XLnet and BERT models implemented using pytorch-transformers library \footnote{\url{https://huggingface.co/transformers/}} and were topped with two dense fully connected layers used as the output classifier.

\end{document}